\documentclass[journal]{IEEEtran}

\ifCLASSOPTIONcompsoc
  \usepackage[nocompress]{cite}
\else
  \usepackage{cite}
\fi
\usepackage[utf8]{inputenc}
\usepackage{amssymb}
\usepackage{booktabs}
\usepackage{amsmath}
\usepackage{multirow}
\usepackage{url}
\usepackage{xcolor}
\usepackage{graphicx}
\usepackage[hidelinks]{hyperref}

\newif\ifanonymous
\ifdefined\anonymousbuild\anonymoustrue\else\anonymousfalse\fi

\newcommand{\anonid}{task-order-cl-4f7b}

\newcommand{\ord}[4]{#1$\to$#2$\to$#3$\to$#4}

\begin{document}
\bstctlcite{IEEEcontrol}

\title{What Is the Influence of Task Order in Continual Learning? An Empirical Study and Order-Aware Evaluation for Motion Forecasting}

\ifanonymous
\author{Anonymous Authors}
\markboth{IEEE Transactions on Neural Networks and Learning Systems,~Vol.~XX, No.~X, 2026}%
{Anonymous: What Is the Influence of Task Order in Continual Learning?}
\else
\author{Zirui Li,~\IEEEmembership{Member,~IEEE}, Yunlong Lin, Haohan Yang, Xiaocong Zhao,~\IEEEmembership{Member,~IEEE}, and Chen Lv,~\IEEEmembership{Senior Member,~IEEE}
\thanks{This work was supported by the Nanyang Technological University (NTU) AI-for-X Postdoctoral Fellowship of National Research Foundation (NRF) under Grant 026374-00006. \textit{(Corresponding author: Chen Lv.)}}
\thanks{Zirui Li and Chen Lv are with the School of Mechanical and Aerospace Engineering, Nanyang Technological University, Singapore 639798 (e-mail: lyuchen@ntu.edu.sg).}
\thanks{Yunlong Lin is with the School of Mechanical Engineering, Beijing Institute of Technology, Beijing 100081, China, and also with the School of Computation, Information and Technology, Technical University of Munich, 85748 Garching, Germany.}
\thanks{Haohan Yang is with the Department of Civil and Environmental Engineering, Samueli School of Engineering, University of California, Los Angeles, CA 90095 USA.}
\thanks{Xiaocong Zhao is with the Key Laboratory of Road and Traffic Engineering, Ministry of Education, Tongji University, Shanghai 201804, China.}
}
\markboth{IEEE Transactions on Neural Networks and Learning Systems,~Vol.~XX, No.~X, 2026}%
{Li \MakeLowercase{\textit{et al.}}: What Is the Influence of Task Order in Continual Learning?}
\fi

\IEEEtitleabstractindextext{%
\begin{abstract}
Aggregate scores for continual learning algorithms rest on a distribution over task orders that is rarely stated. We formulate algorithm comparison as an expectation over a declared order distribution, which separates training variation from order variation, and apply it to motion forecasting with four road-layout tasks from the INTERACTION dataset, all 24 task orders, four algorithms, and three training replicates (288 runs). Task order accounts for 65\% to 91\% of the variation in current-task error and backward transfer, and 90\% to 99\% of this order effect comes from the identity of the starting task: the smallest task cannot be learned from a cold start. Beyond the first stage, the order effect is comparable to training variability. In most orders, the leading algorithm in current-task error and backward transfer is within one replicate standard deviation of the runner-up. Backward transfer depends on the order mainly through its reference term, which weights the first task most heavily, so relative improvement and final error identify different leading algorithms for displacement error. A transition-direction descriptor computed from task statistics tracks the position of the smallest task, and weighting orders by it leaves the algorithm ranking unchanged. These results call for benchmarks that declare their order distribution and report order-resolved, starting-task-resolved, warm-start, and final errors alongside aggregate scores. The code and data are available \href{https://anonymous.4open.science/r/task-order-cl-4f7b/}{here}.
\end{abstract}
\begin{IEEEkeywords}
Benchmark testing, continual learning, evaluation protocol, motion forecasting, task order.
\end{IEEEkeywords}}

\maketitle
\IEEEdisplaynontitleabstractindextext
\IEEEpeerreviewmaketitle

\begin{figure}[!t]
\centering
\includegraphics[width=\columnwidth]{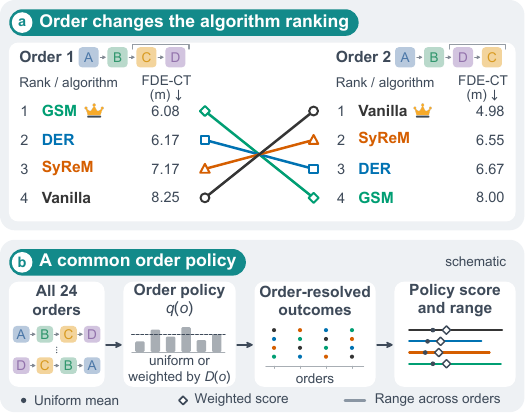}
\caption{Task order and algorithm comparison. (a) Current-task final displacement error (FDE-CT, m), averaged over three replicates, and algorithm ranking for two orders that differ only in their final two tasks. In both orders the leader lies within one pooled replicate standard deviation of the runner-up. (b) A common order policy $q(o)$, uniform or weighted by the transition descriptor $D(o)$, aggregates the order-resolved outcomes of every algorithm into a uniform mean or weighted score, shown with the range across orders.}
\label{fig:teaser}
\end{figure}

\section{Introduction}
\label{sec:introduction}

\IEEEPARstart{S}{elf-driving} vehicles plan their motion around forecasts of where surrounding road users will go. A forecaster trained on urban merges meets highway ramps, intersections, and roundabouts as the fleet expands, and must keep learning in an order no designer controls. Continual learning (CL) provides this capability and is moving into safety-relevant applications. Human-guided CL already uses small amounts of driving guidance to improve self-driving performance in dimensions such as social compliance and rare-case handling~\cite{yang2026humanguided}, and continual forecasting methods adapt motion predictors across changing traffic scenarios~\cite{lin2023rethinking,lin2024continual,lin2025escaping-arxiv,lin2025h2c,li2025dual-ls}. Learning a new task can overwrite earlier knowledge~\cite{mccloskey1989catastrophic,forget,parisi2019continual-lifelong-review}. Replay, regularisation, and gradient constraints counter this forgetting through different retention mechanisms~\cite{kirkpatrick2017EWC,li2017learningLwF,lopez2017gradient-gem,chaudhry2019efficient-agem,buzzega2020dark}, so their outcomes depend on the learning history fixed by the task order. Yet the algorithm to deploy is chosen from benchmark rankings. If a ranking reverses when the same tasks arrive in another sequence, that choice hinges on an arbitrary benchmark order.

Because outcomes depend on the order in which tasks arrive, any aggregate score for a CL algorithm rests on a distribution over task orders. Evaluating continual forecasting under varying data-stream sequences has been recommended~\cite{li2023continual-survey}, yet many protocols leave this distribution implicit, repeating one or a few prescribed orders while varying initialisation or within-task sample order~\cite{buzzega2020dark,lin2025h2c}. Studies of order-dependent forgetting, order-wise performance disparity, ordering principles, and sequence transferability show that task transitions influence retention and transfer~\cite{bell2022effect,yoon2020apd,li2025optimal-order,nguyen2025sequence}, and use this to explain the outcomes of particular orders, quantify the disparity among them, or select favourable ones. How the variation of a metric across all $N!$ orders of $N$ tasks, several algorithms, and repeated training runs divides between order, algorithm, their interaction, and training randomness, and which property of an order carries the order effect, remain open.

\begin{figure*}[!t]
\centering
\includegraphics[width=\textwidth]{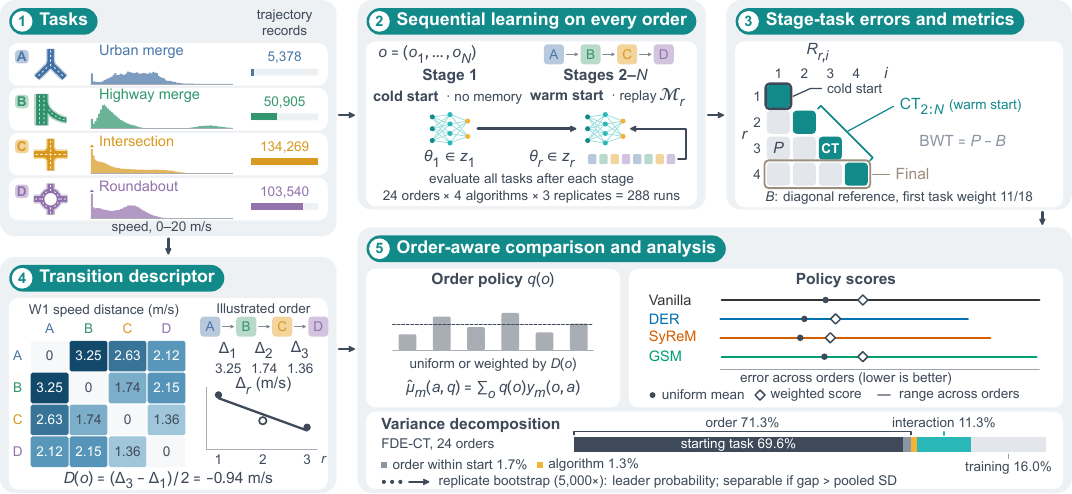}
\caption{Order-aware evaluation workflow. (1) Tasks differ in size and speed distribution. (2) Each algorithm learns every order, starting cold in the first stage. (3) Testing all tasks after each stage gives the stage--task errors $R_{r,i}$, summarised by CT, $\mathrm{CT}_{2:N}$, $\mathrm{BWT}=P-B$, and the final error. (4) W1 speed distances between consecutive tasks, $\Delta_r$, define the descriptor $D(o)$. (5) A common order policy $q(o)$ yields policy scores, and a sum-of-squares decomposition (FDE-CT) and a replicate bootstrap assess them. Networks, policy weights, and comparison marks are schematic.}
\label{fig:pipeline}
\end{figure*}

This paper answers both questions for a fixed task set and tests whether algorithm rankings persist across orders. We formulate algorithm comparison as an expectation over an explicit order policy, with the uniform policy as the equal-order reference, and decompose the sum of squares of each metric into order, algorithm, interaction, and training components, splitting the order term by starting task. A replicate bootstrap attaches a probability to every leading algorithm. A warm-start current-task error and a decomposition of backward transfer into subsequent error and initial reference isolate the first learning stage. A transition-direction descriptor computed from task statistics defines a descriptor-weighted policy, an example of a nonuniform policy shared by all algorithms (Figs.~\ref{fig:teaser}(b) and~\ref{fig:pipeline}).

The experiments use the UQnet endpoint-heatmap backbone~\cite{li2024UQnet} and four road-layout tasks from the INTERACTION dataset: Urban Merge (A), Highway Merge (B), Intersection (C), and Roundabout (D). Vanilla sequential fine-tuning and three memory-based algorithms learn the tasks in all $4!=24$ orders, with three training replicates per order and algorithm (288 runs). The memory-based algorithms are dark experience replay (DER), synergetic memory rehearsal (SyReM), and gradient scenario memory (GSM). In Fig.~\ref{fig:teaser}(a), \ord{A}{B}{C}{D} and \ord{A}{B}{D}{C} differ only in their final two tasks, yet give exactly reversed rankings of the four algorithms by current-task final displacement error. Section~\ref{subsec:order_resolved} traces both rankings to run-to-run variation in the cold-start error, which arises on the shared first task before the two orders diverge.

The main contributions are summarised as follows.
\begin{enumerate}
\item \emph{Formulation and benchmark.} An expectation over an explicit order policy defines a common comparison target, and the law of total variance separates within-order training variation from across-order variation. All 24 orders are enumerated with released per-run results, and the decomposition, adjustment, and bootstrap procedures form a reusable analysis toolkit.
\item \emph{Empirical finding.} Task order accounts for 65\% to 91\% of the variation in current-task error and backward transfer, and 90\% to 99\% of this order effect comes from the identity of the starting task, acting through the cold start of the smallest task. Beyond the first stage, the residual order effect is comparable to training variability. The per-order leader's gap to the runner-up exceeds the pooled replicate standard deviation in only 2 to 11 of the 24 orders for current-task error and backward transfer, against 16 or 17 for final error.
\item \emph{Metric diagnosis.} The first task enters the reference term of stage-averaged backward transfer with weight $11/18$ for four tasks, so a poor first fit simultaneously raises current-task error and drives backward transfer towards large negative values. Relative improvement and absolute final error therefore identify different leading algorithms for displacement error. The transition-direction descriptor tracks the position of the smallest task, and the descriptor-weighted policy leaves the full ranking unchanged.
\end{enumerate}

\section{Related Work}
\label{sec:related}

\subsection{Task Order and Curriculum Effects}
\label{subsec:rw_order}
Curriculum learning investigates the effect of example or task presentation on optimisation~\cite{bengio2009curriculum}. In CL, ordering also affects retention of previous knowledge. Bell and Lawrence~\cite{bell2022effect} examine order-dependent forgetting and task-distance heuristics, and Li and Hiratani~\cite{li2025optimal-order} derive ordering principles in a teacher--student model. Analyses of task similarity and overparameterisation further characterise interference~\cite{goldfarb2024joint}. Nguyen \emph{et al.}~\cite{nguyen2019understanding} relate error rates to the total complexity of a task sequence and to its sequential heterogeneity, the sum of distances between consecutive tasks. The path length $L$ in Section~\ref{subsec:analysis_protocol} is such a sum.

Yoon \emph{et al.}~\cite{yoon2020apd} quantify order-wise performance disparity, establishing order fairness as an evaluation criterion. Measures of sequence transferability relate task transitions to performance using learned predictors~\cite{nguyen2025sequence}. The present analysis uses marginal task statistics to define a shared order policy and asks how much of an observed ranking survives a change of order.

\subsection{Continual Learning for Motion Forecasting}
\label{subsec:rw_forecasting}
Continual forecasting addresses adaptation across changing traffic environments~\cite{li2023continual-survey} through generative replay~\cite{bao2023lifelong,wu2022continual-ped,ma2021continual}, regularisation under distribution shift~\cite{2022-EWC-ped-traj}, and task-free streams without supplied task boundaries~\cite{lin2023rethinking,lin2025h2c,aljundi2019task-free}.

D-GSM uses traffic divergence and scenario-organised memory to constrain gradient updates~\cite{lin2024continual}, H2C combines diversity-based selection with equiprobable replay~\cite{lin2025h2c}, SyReM integrates a memory-loss constraint with gradient-based rehearsal selection~\cite{lin2025escaping-arxiv}, and Dual-LS coordinates long- and short-term knowledge through complementary rehearsal~\cite{li2025dual-ls}. These studies evaluate each method along one or a few task orders, so the dependence of the resulting ranking on the order is still unexamined.

\subsection{Evaluation Protocols for Continual Learning}
\label{subsec:rw_eval}
Average performance over all tasks, current-task performance, and backward transfer are the standard quantities of continual learning evaluation~\cite{lopez2017gradient-gem,lange2022-CL-survey}, and the appropriate protocol differs between task-, domain-, and class-incremental settings~\cite{van2022three-incremental}. Early protocol studies showed that reported forgetting depends on the evaluation design, including the task sequences used~\cite{kemker2018measuring,farquhar2018robust}, and proposed metric sets beyond final accuracy~\cite{diaz2018dont}. Hyperparameter selection on the evaluation scenario can bias comparisons~\cite{cha2024hyperparameters}, and unequal update costs can alter conclusions about online CL~\cite{ghunaim2023real-ocl}. Recent benchmarks treat the order either as an adversarial choice to be searched~\cite{cldyb2025} or as a sampled factor whose uncertainty is reported~\cite{medclbench2026}.

Order coverage is therefore a protocol choice. We distinguish repeated training within an order from evaluation across orders, separate the starting task from the remaining order, and assess backward transfer alongside initial, warm-start, and final error.

\section{Problem Formulation}
\label{sec:formulation}
\subsection{Sequential Learning Conditioned on Tasks}
\label{subsec:cl_preliminaries}
Let $[N]=\{1,\ldots,N\}$ index a task set $\mathcal T$. Task $j$ is specified by a distribution $\mathcal D_j$ on $\mathcal X\times\mathcal Y$ and a finite training set $\mathcal S_j=\{(\mathbf X_{jk},\mathbf Y_{jk})\}_{k=1}^{n_j}$. The tasks share the input and output spaces, while their data distributions may differ. An order (task sequence) $o=(o_1,\ldots,o_N)$ belongs to the permutation group $\mathfrak S_N$, with $|\mathfrak S_N|=N!$. Evaluation uses a common subset $\mathcal O\subseteq\mathfrak S_N$ for every algorithm.

For algorithm $a\in\mathcal A$, let $z_r^{(o,a,s)}$ denote its state after stage $r$ in training replicate $s$, comprising model parameters, optimiser state, and algorithm-specific persistent information such as the replay memory $\mathcal M_r^{(o,a,s)}\subseteq z_r^{(o,a,s)}$. Sequential learning is represented by
\begin{equation}
z_r^{(o,a,s)}=\mathcal U_a\bigl(z_{r-1}^{(o,a,s)},\mathcal S_{o_r},\xi_r^{(o,a,s)}\bigr),
\label{eq:update}
\end{equation}
where $\mathcal U_a$ is the update operator and $\xi_r^{(o,a,s)}$ denotes the stochastic inputs for that run and stage. The initial state is likewise run-specific. The memory satisfies $|\mathcal M_r^{(o,a,s)}|\le B_a$ for a record capacity $B_a$, a record may include a stored model output, and historical training examples are accessible only through this memory. Task boundaries are provided, and all tasks share one prediction head.

\subsection{Risk and Knowledge Retention}
\label{subsec:cl_objective}
For loss $\ell$ and predictor $f_{\boldsymbol\theta}$, the risk on task $j$ is
\begin{equation}
\mathcal R_j(\boldsymbol\theta)=\mathbb E_{(\mathbf X,\mathbf Y)\sim\mathcal D_j}
 [\ell(f_{\boldsymbol\theta}(\mathbf X),\mathbf Y)].
\label{eq:task_risk}
\end{equation}
A reference objective for stage $r>1$ is
\begin{equation}
\boldsymbol\theta_r^\star\in\arg\min_{\boldsymbol\theta}
\left\{\mathcal R_{o_r}(\boldsymbol\theta)+
\frac{\lambda_r}{r-1}\sum_{i<r}\mathcal R_{o_i}(\boldsymbol\theta)\right\},
\label{eq:dilemma}
\end{equation}
where $\lambda_r\ge0$ balances current-task fit and past-task risk. Stage $1$ contains only the current-task term, so the first stage is single-task learning from the initial state. Replay, regularisation, and gradient constraints approximate the historical risk through different mechanisms~\cite{kirkpatrick2017EWC,li2017learningLwF,buzzega2020dark,lopez2017gradient-gem}. Absolute historical risk and its change since initial learning are distinct quantities, which final error and backward transfer measure separately.

\subsection{Evaluation over Task Orders}
\label{subsec:order_estimand}
Let $Z_m(o,a,\xi)$ be the scalar outcome of a complete run under metric $m$, where $\xi$ collects initialisation and subsequent training randomness for the specified order and algorithm. For an order policy $q$ satisfying $q(o)\ge0$ and $\sum_{o\in\mathcal O}q(o)=1$, the comparison target is
\begin{equation}
\mu_m(a,q)=\sum_{o\in\mathcal O}q(o)\,
\mathbb E_{\xi}[Z_m(o,a,\xi)].
\label{eq:population_policy}
\end{equation}
For $S$ training replicates, define $y_m(o,a)=S^{-1}\sum_{s=1}^{S}Z_m(o,a,\xi^{(o,a,s)})$. The empirical policy score is
\begin{equation}
\widehat\mu_m(a,q)=\sum_{o\in\mathcal O}q(o)y_m(o,a).
\label{eq:policy}
\end{equation}
All algorithms share $\mathcal O$, $q$, and the metric definition. For fixed $a$, $m$, and $q$, assume $Z_m$ has a finite second moment and draw $o\sim q$. The law of total variance separates the two sources of variability:
\begin{align}
\operatorname{Var}(Z_m)
&=\mathbb E_{o\sim q}[\operatorname{Var}_{\xi}(Z_m\mid o)]\nonumber\\
&\quad+\operatorname{Var}_{o\sim q}(\mathbb E_{\xi}[Z_m\mid o]).
\label{eq:variation_axes}
\end{align}
The first term describes stochastic training variation within orders, and the second describes variation in expected performance across orders. Repeating a single order estimates only the first component, so order-aware evaluation combines an explicit order policy with order-resolved outcomes.

For algorithms $a$ and $b$, let $\delta_m(o,a,b)=y_m(o,a)-y_m(o,b)$ be the measured difference. Equation~\eqref{eq:policy} gives
\begin{equation}
\widehat\mu_m(a,q)-\widehat\mu_m(b,q)=\sum_{o\in\mathcal O}q(o)\delta_m(o,a,b).
\label{eq:policy_gap}
\end{equation}
A common sign of $\delta_m$ across orders preserves the weak ranking under every policy supported on $\mathcal O$, whereas opposite signs permit a reversal under policies concentrating on different orders. Order-wise differences thus distinguish uniform dominance from dependence on the evaluation distribution. Section~\ref{subsec:analysis_protocol} quantifies how replicate variability affects such comparisons.

\section{Order-Aware Evaluation}
\label{sec:order_aware}
\subsection{Distances Between Task Distributions}
\label{subsec:dist_shift}
We describe each task through empirical distributions of target-agent speed and acceleration. Speed is the norm of the two velocity components at each of nine historical time steps, and acceleration is the adjacent speed difference divided by 0.1~s, retaining values in $[-5,5]$~m/s$^2$. Observations are pooled over time steps and trajectories within each task.

The candidate measures probe complementary aspects of distribution shift. Wasserstein-1 (W1) quantifies displacement of probability mass along the physical feature axis~\cite{peyre2019computational}. Kullback--Leibler (KL) divergence measures directed probability mismatch, whereas Jensen--Shannon (JS) divergence provides a symmetric, bounded comparison of the same histograms~\cite{lin1991divergence}. Maximum mean discrepancy (MMD) compares kernel mean embeddings without histogram discretisation~\cite{gretton2012kernel}. Applying all four measures to speed holds the feature representation fixed while varying the notion of discrepancy, whereas applying W1 additionally to acceleration examines a different aspect of motion with the same distance functional.

For feature $f\in\{\text{speed},\text{acceleration}\}$, let $F_i^f$ denote its empirical cumulative distribution function in task $i$. We compute
\begin{equation}
d^{\rm W}_{ij}(f)=\int_{\mathbb R}|F_i^f(t)-F_j^f(t)|\,\mathrm{d}t,
\label{eq:symmetry}
\end{equation}
using every retained observation. The distance is expressed in m/s for speed and m/s$^2$ for acceleration.

KL and JS divergences are computed on $H=100$ shared equal-width speed histograms with additive smoothing, and MMD on a fixed standardised sample of $2{,}000$ speed observations per task with a Gaussian kernel (Appendix~\ref{subsec:appendix_distance}). W1, JS, and MMD are symmetric, whereas KL is directed.

The primary distance is W1 on speed, which has an interpretable physical scale. It was chosen in an exploratory scan of 20 distance representations and 15 aggregation operators against the CL outcomes (Appendix~\ref{subsec:appendix_scan}), from which the five candidates above are reported.

\subsection{A Descriptor of Transition Direction}
\label{subsec:selecting_distance}
For order $o$, let $\Delta_r(o)=d_{o_r,o_{r+1}}$ be the distance at the $r$th transition. The descriptor is its least-squares slope against transition position:
\begin{equation}
D(o)=\frac{\sum_{r=1}^{N-1}(r-\bar r)\Delta_r(o)}
{\sum_{r=1}^{N-1}(r-\bar r)^2},\qquad \bar r=N/2.
\label{eq:Do}
\end{equation}
It is defined for $N\ge3$. For the four-task experiment this reduces exactly to
\begin{equation}
D(o)=\frac{d_{o_3,o_4}-d_{o_1,o_2}}{2}.
\label{eq:four_task_D}
\end{equation}
The middle transition has zero coefficient, and a positive value means that the final transition is larger than the first. We therefore call $D$ a transition-direction descriptor.

For symmetric distances, $D(o^{\rm rev})=-D(o)$, where $o^{\rm rev}$ reverses the order. There are only three partitions of four tasks into two unordered pairs, and each partition gives two signed contrasts, so~\eqref{eq:four_task_D} takes at most six distinct values. Swapping the tasks within either endpoint pair leaves $D$ unchanged, so each value is shared by four orders. The slope is equivariant to a common positive affine transformation of the distances, $d'_{ij}=a d_{ij}+b$ with $a>0$ giving $D'(o)=aD(o)$, so its rank and the min--max weights below are unchanged by a common linear rescaling.

In this task set, the first and last transitions of any order form one of the three pair partitions, and in every partition the pair containing task A has the larger speed distance ($d_{AB}=3.25>d_{CD}=1.36$, $d_{AC}=2.63>d_{BD}=2.15$, and $d_{AD}=2.12>d_{BC}=1.74$~m/s). The sign of $D$ in~\eqref{eq:four_task_D} is therefore set by the position of task A: $D(o)<0$ exactly when A occupies position 1 or 2, and the Spearman correlation between $D$ and the position of A is 0.79. Section~\ref{subsubsec:dvscl} accordingly examines associations with both.

\subsection{Uniform and Descriptor-Weighted Policies}
\label{subsec:weighted_metric}
The uniform policy $q(o)=1/|\mathcal O|$ assigns equal probability to each evaluated order and gives the equal-order reference
\begin{equation}
\bar y_m(a)=\frac1{|\mathcal O|}\sum_{o\in\mathcal O}y_m(o,a).
\label{eq:uniform}
\end{equation}
We report it together with the order-resolved means, ranks, and replicate spread.

The descriptor-weighted policy rescales the descriptor to the unit interval, with $D_{\min}=\min_{o\in\mathcal O}D(o)$ and $D_{\max}=\max_{o\in\mathcal O}D(o)$:
\begin{equation}
\widetilde D(o)=\frac{D(o)-D_{\min}}{D_{\max}-D_{\min}}.
\label{eq:Dtilde}
\end{equation}
If $D$ is constant, the weights are uniform and the weighted result equals~\eqref{eq:uniform}. Otherwise, following the directions of association in Section~\ref{subsubsec:dvscl}, the common weights for the current-task error (CT) and backward transfer (BWT), both defined in Section~\ref{sec:metrics}, are
\begin{equation}
w_{\rm CT}(o)=1-\widetilde D(o),\qquad
w_{\rm BWT}(o)=\widetilde D(o),
\label{eq:weights}
\end{equation}
and, for each metric separately,
\begin{equation}
\bar y_m^{\,w}(a)=\frac{\sum_o w_m(o)y_m(o,a)}{\sum_o w_m(o)}.
\label{eq:weighted_metric}
\end{equation}
The normalised weights define $q_m(o)=w_m(o)/\sum_{o'}w_m(o')$ in~\eqref{eq:policy}. Because the complete order set contains the reversal of every order, a nonconstant symmetric-distance descriptor gives $D_{\min}=-D_{\max}$, so both weight vectors sum to $|\mathcal O|/2$. The policy emphasises the descriptor regions associated with larger current-task errors and less favourable backward transfer.

\subsection{Analysis Protocol}
\label{subsec:analysis_protocol}
The decomposition and the bootstrap operate on individual runs, and the other analyses use the replicate means $y_m(o,a)$.

\emph{Decomposition.} For each metric, the total centred sum of squares over all runs is split into order, algorithm, order--algorithm interaction, and within-cell training components (Appendix~\ref{subsec:appendix_decomp}). The order component is further split into starting-task identity and order within starting task, and the decomposition is repeated for the 18 orders that do not start with task A.

\emph{Warm-start error.} To separate the first stage, which involves no previous task, from the continual part of a run, the current-task error is also reported without its first term (Section~\ref{sec:metrics}).

\emph{Associations.} Spearman correlations with the 16 algorithm--metric series use average ranks for ties. Starting-task dependence is assessed by excluding A-first orders and by correlating ranked variables after removing starting-task group means. The same residualisation with position-of-A group means isolates what $D$ adds beyond the position of task A.

\emph{Replicate bootstrap.} The three replicates of every order--algorithm cell are resampled with replacement (5,000 draws, fixed seed), and the order means, policy scores, and per-order leaders are recomputed. This yields the probability that each algorithm attains the smallest policy score and an interval for its number of per-order wins. A per-order leader is called separable when its gap to the runner-up exceeds the pooled sample standard deviation of the two algorithms. Because the 24 orders form the complete order set, order sampling adds no uncertainty, and the resampling reflects training variation alone (Appendix~\ref{subsec:appendix_decomp}).

\emph{Paths.} We also compare the orders with the largest and smallest total transition distance,
\begin{equation}
L(o)=\sum_{r=1}^{N-1}d_{o_r,o_{r+1}},
\label{eq:path_length}
\end{equation}
within each fixed starting task. Path length and slope are different summaries: $L$ uses all three transitions, whereas $D$ omits the middle one.

\section{Experimental Setup}
\label{sec:settings}
\subsection{Motion Forecasting Tasks and Backbone}
\label{subsec:dataset}
Each forecasting task comprises trajectory records from one category of road layout in INTERACTION~\cite{zhan2019interaction-dataset}. Twelve location subsets are grouped into four tasks. Task A, \emph{Urban Merge}, covers lane convergence and zipper merging at the German MT site. Task B, \emph{Highway Merge}, covers ramp entry and lane convergence at the Chinese ZS highway site (archived subsets ZS0 and ZS2). Task C, \emph{Intersection}, covers crossing and turning interactions at the US subsets MA, GL, EP0, and EP1. Task D, \emph{Roundabout}, covers entry, circulation, and exit at EP, SR, and FT in the US, OF in Germany, and LN in China. The task arrays contain 5,378, 50,905, 134,269, and 103,540 trajectory records, so task A is 10 to 25 times smaller than the others.

\begin{figure}[!t]
\centering
\includegraphics[width=\columnwidth]{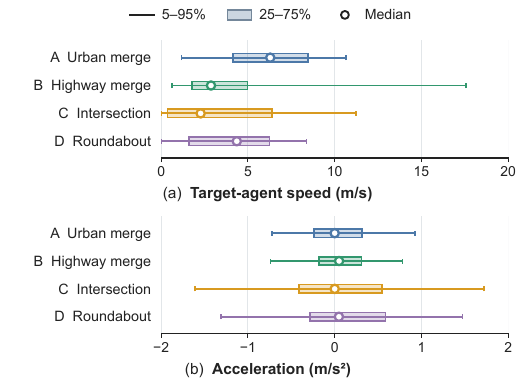}
\caption{Task feature distributions. Dots, bands, and whiskers indicate medians, interquartile ranges, and 5th--95th percentiles. Acceleration is restricted to $[-5,5]$~m/s$^2$.}
\label{fig:task_distributions}
\end{figure}

Fig.~\ref{fig:task_distributions} compares the locations and spreads of the empirical target-agent feature distributions. Urban merging has the highest median speed (6.29~m/s), whereas highway merging combines a lower median (2.87~m/s), reflecting queuing on the ramp, with the largest 95th percentile (17.56~m/s). Intersections have the lowest median speed (2.27~m/s), and intersections and roundabouts have the wider acceleration interquartile ranges.

The forecasting input is $\mathbf X=(\mathbf X_{\rm TA},\mathbf X_{\rm SA},\mathbf M)$, containing the target-agent history, surrounding-agent histories, and map context. The ground-truth output $\mathbf Y\in\mathbb R^2$ is the target endpoint. UQnet~\cite{li2024UQnet} combines agent and map encoders with a spatial prediction head to produce a heatmap over future endpoints, from which the predictor samples $W$ candidate endpoints $\{\hat{\mathbf Y}_w\}_{w=1}^W$. The training loss follows the heatmap-based forecasting formulation in the backbone work.

\begin{figure}[!t]
\centering
\includegraphics[width=\columnwidth]{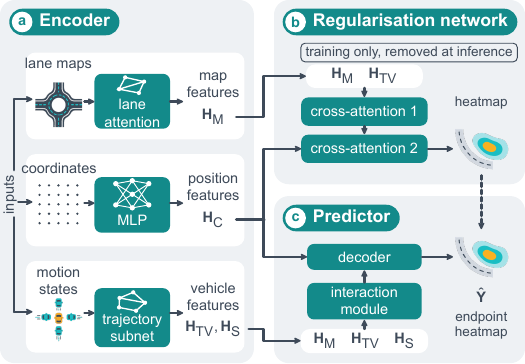}
\caption{UQnet backbone (adapted from~\cite{li2024UQnet}): (a) encoder, (b) regularisation network, used during training and removed at inference, and (c) predictor, which decodes fused map and vehicle features with position features into an endpoint heatmap.}
\label{fig:uqnet}
\end{figure}

Fig.~\ref{fig:uqnet} details the encoder and prediction branches. Lane attention encodes the map into $\mathbf H_{\rm M}$, a multilayer perceptron embeds spatial coordinates as $\mathbf H_{\rm C}$, and a trajectory subnet extracts target-vehicle and surrounding-agent features, $\mathbf H_{\rm TV}$ and $\mathbf H_{\rm S}$. The interaction module fuses vehicle and map features, which the decoder combines with coordinate features to predict an endpoint heatmap. The regularisation network applies two cross-attention blocks to map, target-vehicle, and coordinate features during training and is removed at inference~\cite{li2024UQnet}. All compared algorithms use this predictor, and evaluation concerns the predicted endpoints.

\subsection{Algorithms and Experimental Coverage}
\label{subsec:algorithms}
We compare four algorithms under the common prediction architecture (Table~\ref{tab:algorithm_coverage}). \emph{Vanilla} optimises the current task without memory and provides the reference for unconstrained adaptation. \emph{DER} (dark experience replay)~\cite{buzzega2020dark} rehearses stored inputs with output matching to preserve previous predictions, in a forecasting adaptation of the original classification method. \emph{SyReM} (synergetic memory rehearsal)~\cite{lin2025escaping-arxiv} combines gradient-based rehearsal selection with an average-memory loss constraint. \emph{GSM}~\cite{lin2024continual} is the equal-allocation variant of dynamic gradient scenario memory (D-GSM), with scenario-organised memory (a scenario is one of the four tasks here) and gradient constraints derived from previous scenarios. The selection spans output distillation, selective constrained rehearsal, and scenario-wise constraints, and all memory-based algorithms use a capacity of 2,000 records.

\begin{table}[!t]
\centering
\caption{Published mechanisms of the compared algorithms. Stored examples serve output matching, rehearsal, or gradient constraints, and scenario partition allocates memory by past task.}
\label{tab:algorithm_coverage}
\footnotesize
\setlength{\tabcolsep}{2pt}
\renewcommand{\arraystretch}{1.0}
\begin{tabular*}{\columnwidth}{@{\extracolsep{\fill}}lccccc@{}}
\toprule
Algorithm & \shortstack{Stored\\examples} & \shortstack{Output\\matching} & \shortstack{Scenario\\partition} & \shortstack{Gradient\\constraint} & \shortstack{Gradient-based\\selection}\\
\midrule
Vanilla & $\times$ & $\times$ & $\times$ & $\times$ & $\times$\\
DER & $\checkmark$ & $\checkmark$ & $\times$ & $\times$ & $\times$\\
SyReM & $\checkmark$ & $\times$ & $\times$ & $\checkmark$ & $\checkmark$\\
GSM & $\checkmark$ & $\times$ & $\checkmark$ & $\checkmark$ & $\times$\\
\bottomrule
\end{tabular*}
\end{table}

The experiment evaluates all $4!=24$ orders of $N=4$ tasks, with four algorithms and three training replicates per combination (288 runs). Testing every task after each stage yields 1,152 stage logs and 4,608 stage--task records. The evaluation data and analysis code are available online.\footnote{Code and data: \url{https://anonymous.4open.science/r/task-order-cl-4f7b/}.}

\subsection{Forecasting and Continual Learning Metrics}
\label{sec:metrics}
Minimum final displacement error (FDE) measures the distance from the ground-truth endpoint to the closest predicted candidate:
\begin{equation}
\operatorname{FDE}=\min_{1\le w\le W}\|\hat{\mathbf Y}_w-\mathbf Y\|_2.
\label{eq:fde}
\end{equation}
Miss rate (MR) reports the percentage of evaluated cases whose closest candidate lies beyond the endpoint tolerance of the evaluation protocol.

For either error metric, let $E_{r,j}^{(o,a,s)}$ denote the result on task identity $j$ after stage $r$. We reorder columns by presentation position and define
\begin{equation}
R_{r,i}^{(o,a,s)}=E_{r,o_i}^{(o,a,s)}.
\label{eq:R}
\end{equation}
Thus $R_{r,r}$ always evaluates the task just learned, even when its identity is not $r$. Suppressing the superscripts, the summaries are
\begin{align}
\mathrm{CT}&=\frac1N\sum_{r=1}^{N}R_{r,r},\label{eq:CT}\\
\mathrm{BWT}&=\frac1{N-1}\sum_{r=2}^{N}\frac1{r-1}
\sum_{i=1}^{r-1}(R_{r,i}-R_{i,i}),\label{eq:BWT}\\
\mathrm{Final}&=\frac1N\sum_{i=1}^{N}R_{N,i}.\label{eq:final}
\end{align}
The current-task error CT measures error upon initial learning of each task. The backward transfer BWT averages subsequent error changes over stages: positive values indicate degradation and negative values improvement relative to the initial fit, reversing the accuracy-based sign convention. FDE is in metres, MR in percent, and MR-BWT in percentage points. Lower CT and Final indicate better absolute predictions, and lower BWT indicates greater improvement relative to the initial fit.

The first term of CT, $R_{1,1}$, is the error of single-task learning from the initial state and involves no previous task. We therefore also report the warm-start current-task error
\begin{equation}
\mathrm{CT}_{2:N}=\frac1{N-1}\sum_{r=2}^{N}R_{r,r},
\label{eq:CT_warm}
\end{equation}
so that $\mathrm{CT}=[R_{1,1}+(N-1)\,\mathrm{CT}_{2:N}]/N$. We write FDE-CT, MR-BWT, and so on for the metric--summary combinations.

\section{Experimental Results}
\label{sec:results}
\subsection{Order-Resolved Algorithm Comparisons}
\label{subsec:order_resolved}
Tables~\ref{tab:main_results_plasticity} and~\ref{tab:main_results_stability} report all orders. Each entry is a mean and sample standard deviation across three replicates, and bold marks the per-order minima that are separable from the runner-up.

\begin{table}[!t]
\centering
\caption{Percentage of the total centred sum of squares per component. The order term is split into starting-task identity (Start) and order within starting task, and the last three columns give the 18 orders that do not start with A.}
\label{tab:decomposition}
\setlength{\tabcolsep}{1.8pt}
\footnotesize
\begin{tabular*}{\columnwidth}{@{\extracolsep{\fill}}lrrrrrrrr@{}}
\toprule
 & \multicolumn{5}{c}{All 24 orders (\%)} & \multicolumn{3}{c}{Excl.\ A-first (\%)} \\
\cmidrule(lr){2-6}\cmidrule(lr){7-9}
Metric & Start & \shortstack{Order$\mid$\\start} & Alg. & \shortstack{Order$\times$\\alg.} & Within & Order & Alg. & Within \\
\midrule
FDE-CT & $69.6$ & $1.7$ & $1.3$ & $11.3$ & $16.0$ & $30.6$ & $2.3$ & $50.8$ \\
MR-CT & $58.3$ & $6.3$ & $1.7$ & $10.3$ & $23.4$ & $24.5$ & $2.4$ & $51.2$ \\
FDE-BWT & $75.7$ & $0.5$ & $0.9$ & $9.7$ & $13.1$ & $19.3$ & $17.4$ & $45.9$ \\
MR-BWT & $89.8$ & $1.5$ & $3.4$ & $1.3$ & $4.0$ & $39.5$ & $26.5$ & $25.6$ \\
\midrule
FDE-CT$_{2:N}$ & $4.9$ & $20.7$ & $2.3$ & $20.2$ & $52.0$ & $24.6$ & $1.8$ & $53.1$ \\
MR-CT$_{2:N}$ & $7.9$ & $15.9$ & $2.8$ & $24.3$ & $49.2$ & $21.3$ & $2.1$ & $51.2$ \\
\bottomrule
\end{tabular*}
\end{table}

Table~\ref{tab:decomposition} decomposes the total centred sum of squares. Task order accounts for 71.3\%, 64.6\%, 76.3\%, and 91.3\% in FDE-CT, MR-CT, FDE-BWT, and MR-BWT, against 1.3\%, 1.7\%, 0.9\%, and 3.4\% for the algorithm and 11.3\%, 10.3\%, 9.7\%, and 1.3\% for their interaction. Starting-task identity alone accounts for 69.6\%, 58.3\%, 75.7\%, and 89.8\%, which is 90\% to 99\% of the order term. Among the 18 orders that do not start with A, the order term falls to 30.6\%, 24.5\%, 19.3\%, and 39.5\%. Within-cell training variation then becomes the largest component for FDE-CT, MR-CT, and FDE-BWT (50.8\%, 51.2\%, and 45.9\%), whereas for MR-BWT the order term remains largest and the algorithm term (26.5\%) matches within-cell variation (25.6\%). Relative to the share expected from within-cell noise alone (5.8\%, 8.4\%, 4.7\%, and 1.4\%, derived in Appendix~\ref{subsec:appendix_decomp}), the interaction is clearly elevated for FDE-CT and FDE-BWT.

For FDE-CT, the algorithm-averaged order means range from 0.97 to 6.92~m, a spread of 5.95~m. The median gap between the best and worst algorithm within an order is 0.87~m, and the gap between the largest and smallest uniform-order algorithm means is 0.62~m. Changing the order thus moves FDE-CT far more than changing the algorithm, and most of this effect is the identity of the first task.

Every algorithm has the smallest replicate-mean FDE-CT on at least three orders: Vanilla on 3, DER on 9, SyReM on 9, and GSM on 3, with bootstrap intervals of $[1,8]$, $[5,12]$, $[5,12]$, and $[1,6]$. All four also lead on at least one MR-CT order (2, 8, 11, and 3), whereas DER never attains a per-order minimum of either BWT metric. Fig.~\ref{fig:teaser}(a) illustrates such a reversal: \ord{A}{B}{C}{D} and \ord{A}{B}{D}{C} differ only in their final two tasks, and the FDE-CT ranking of the first (GSM, DER, SyReM, Vanilla) is exactly reversed in the second. The replicates resolve the per-order leader from the runner-up in 4, 2, 10, and 11 of the 24 orders for FDE-CT, MR-CT, FDE-BWT, and MR-BWT, and the separable FDE-BWT and MR-BWT leaders are nearly all SyReM (9 of 10 and 10 of 11). In the two orders of Fig.~\ref{fig:teaser}(a) the leaders lie within 0.63 pooled standard deviations of the runners-up. The clearest separable FDE-CT reversal is between \ord{A}{C}{B}{D}, where SyReM leads with 3.31~m, and \ord{A}{C}{D}{B}, where DER leads with 3.48~m, with gaps of 3.65 and 1.10 pooled standard deviations. All four of these orders start with task A, and their rankings largely follow the first-stage error on A, for example 10.7~m for SyReM against 21.5 to 24.8~m for the other algorithms on \ord{A}{C}{B}{D}, whereas the warm-start errors of the four algorithms lie between 0.9 and 1.8~m. Because the first stage is the same for every A-first order, these reversals arise mainly from run-to-run variation in that stage, which the next subsection examines.

\begin{table*}[!t]
\centering
\caption{Current-task errors by task order (mean $\pm$ sample standard deviation over three replicates). FDE is in metres and MR in percent. Bold: per-order minimum separable from the runner-up (Section~\ref{subsec:analysis_protocol}).}
\label{tab:main_results_plasticity}
\setlength{\tabcolsep}{3.5pt}
\renewcommand{\arraystretch}{1.0}
\fontsize{7}{7.7}\selectfont
\begin{tabular*}{\textwidth}{@{\extracolsep{\fill}}lcccccccc}
\toprule
Order & \multicolumn{2}{c}{Vanilla} & \multicolumn{2}{c}{DER} & \multicolumn{2}{c}{SyReM} & \multicolumn{2}{c}{GSM} \\
\cmidrule(lr){2-3}\cmidrule(lr){4-5}\cmidrule(lr){6-7}\cmidrule(lr){8-9}
 & FDE-CT & MR-CT & FDE-CT & MR-CT & FDE-CT & MR-CT & FDE-CT & MR-CT \\
\midrule
A$\to$B$\to$C$\to$D & $8.25_{\pm0.46}$ & $32.83_{\pm2.59}$ & $6.17_{\pm2.13}$ & $31.75_{\pm1.97}$ & $7.17_{\pm3.71}$ & $31.74_{\pm1.65}$ & $6.08_{\pm1.36}$ & $31.84_{\pm0.56}$ \\
A$\to$B$\to$D$\to$C & $4.98_{\pm2.49}$ & $30.83_{\pm4.43}$ & $6.67_{\pm2.39}$ & $32.20_{\pm1.14}$ & $6.55_{\pm2.48}$ & $28.76_{\pm2.27}$ & $8.00_{\pm3.49}$ & $34.38_{\pm0.52}$ \\
A$\to$C$\to$B$\to$D & $7.33_{\pm3.01}$ & $33.28_{\pm4.02}$ & $6.69_{\pm2.57}$ & $29.44_{\pm1.99}$ & {\boldmath$3.31_{\pm0.31}$} & {\boldmath$25.33_{\pm0.67}$} & $6.23_{\pm1.09}$ & $31.86_{\pm2.69}$ \\
A$\to$C$\to$D$\to$B & $9.26_{\pm1.95}$ & $29.96_{\pm1.81}$ & {\boldmath$3.48_{\pm1.02}$} & $28.85_{\pm3.15}$ & $4.71_{\pm1.20}$ & $27.92_{\pm4.05}$ & $5.00_{\pm1.84}$ & $30.21_{\pm1.17}$ \\
A$\to$D$\to$B$\to$C & $9.20_{\pm1.98}$ & $31.51_{\pm2.50}$ & $3.38_{\pm0.70}$ & $28.18_{\pm4.41}$ & $3.90_{\pm1.97}$ & $26.08_{\pm6.67}$ & $9.17_{\pm2.11}$ & $35.83_{\pm0.44}$ \\
A$\to$D$\to$C$\to$B & $5.92_{\pm2.54}$ & $26.34_{\pm4.40}$ & $5.10_{\pm3.73}$ & $25.08_{\pm1.65}$ & $6.53_{\pm2.44}$ & $29.38_{\pm0.54}$ & $6.97_{\pm0.34}$ & $32.37_{\pm2.82}$ \\
B$\to$A$\to$C$\to$D & $2.40_{\pm1.13}$ & $21.10_{\pm8.23}$ & $2.97_{\pm1.77}$ & $21.92_{\pm13.62}$ & $2.16_{\pm0.20}$ & $21.02_{\pm3.63}$ & $2.66_{\pm1.38}$ & $21.34_{\pm10.22}$ \\
B$\to$A$\to$D$\to$C & $2.13_{\pm0.63}$ & $20.75_{\pm6.81}$ & $2.09_{\pm0.65}$ & $19.11_{\pm8.98}$ & $1.61_{\pm0.48}$ & $15.98_{\pm5.36}$ & $2.70_{\pm1.27}$ & $23.88_{\pm7.66}$ \\
B$\to$C$\to$A$\to$D & $1.45_{\pm0.37}$ & $12.31_{\pm4.21}$ & $2.14_{\pm1.02}$ & $22.57_{\pm13.92}$ & $1.21_{\pm0.55}$ & $8.46_{\pm4.87}$ & $1.52_{\pm0.32}$ & $13.24_{\pm4.24}$ \\
B$\to$C$\to$D$\to$A & $1.21_{\pm0.32}$ & $9.16_{\pm1.51}$ & $1.12_{\pm0.30}$ & $7.34_{\pm2.98}$ & $1.25_{\pm0.22}$ & $10.65_{\pm2.05}$ & $1.72_{\pm0.57}$ & $16.51_{\pm9.91}$ \\
B$\to$D$\to$A$\to$C & $1.62_{\pm0.19}$ & $15.08_{\pm3.63}$ & $1.75_{\pm0.41}$ & $13.61_{\pm1.82}$ & $1.48_{\pm0.28}$ & $15.67_{\pm7.44}$ & $1.80_{\pm0.50}$ & $16.02_{\pm7.99}$ \\
B$\to$D$\to$C$\to$A & $1.53_{\pm0.53}$ & $11.71_{\pm3.11}$ & $1.33_{\pm0.54}$ & $12.87_{\pm7.73}$ & $1.63_{\pm0.48}$ & $12.85_{\pm3.69}$ & $1.84_{\pm0.19}$ & $17.31_{\pm3.29}$ \\
C$\to$A$\to$B$\to$D & $1.27_{\pm0.28}$ & $14.74_{\pm5.90}$ & $1.13_{\pm0.25}$ & $10.49_{\pm4.61}$ & $1.28_{\pm0.48}$ & $14.34_{\pm9.44}$ & $1.01_{\pm0.20}$ & $7.81_{\pm3.57}$ \\
C$\to$A$\to$D$\to$B & $1.66_{\pm0.42}$ & $22.08_{\pm6.90}$ & $1.07_{\pm0.35}$ & $9.75_{\pm7.29}$ & $1.15_{\pm0.23}$ & $11.32_{\pm5.51}$ & $0.97_{\pm0.15}$ & $7.47_{\pm3.60}$ \\
C$\to$B$\to$A$\to$D & $1.07_{\pm0.10}$ & $9.64_{\pm1.13}$ & $1.07_{\pm0.30}$ & $6.78_{\pm2.11}$ & $0.94_{\pm0.08}$ & $7.34_{\pm2.35}$ & $1.58_{\pm0.36}$ & $17.08_{\pm8.05}$ \\
C$\to$B$\to$D$\to$A & $1.02_{\pm0.11}$ & $8.37_{\pm1.34}$ & $1.12_{\pm0.37}$ & $11.67_{\pm8.13}$ & $1.00_{\pm0.15}$ & $8.76_{\pm3.74}$ & $1.05_{\pm0.26}$ & $7.85_{\pm3.42}$ \\
C$\to$D$\to$A$\to$B & $1.32_{\pm0.79}$ & $12.34_{\pm10.76}$ & {\boldmath$0.91_{\pm0.13}$} & $6.68_{\pm3.23}$ & $1.22_{\pm0.63}$ & $9.62_{\pm8.52}$ & $1.16_{\pm0.20}$ & $11.07_{\pm2.88}$ \\
C$\to$D$\to$B$\to$A & $1.05_{\pm0.34}$ & $7.88_{\pm3.92}$ & $0.95_{\pm0.03}$ & $7.58_{\pm0.53}$ & $0.89_{\pm0.09}$ & {\boldmath$5.91_{\pm2.21}$} & $0.99_{\pm0.18}$ & $8.20_{\pm4.24}$ \\
D$\to$A$\to$B$\to$C & $1.76_{\pm0.93}$ & $18.29_{\pm13.10}$ & {\boldmath$1.08_{\pm0.20}$} & $10.46_{\pm4.87}$ & $1.84_{\pm1.25}$ & $17.54_{\pm11.58}$ & $2.04_{\pm0.82}$ & $21.88_{\pm10.08}$ \\
D$\to$A$\to$C$\to$B & $1.75_{\pm0.12}$ & $23.72_{\pm5.06}$ & $1.14_{\pm0.17}$ & $13.03_{\pm4.88}$ & $1.37_{\pm0.59}$ & $12.03_{\pm4.42}$ & $2.92_{\pm2.14}$ & $23.69_{\pm7.29}$ \\
D$\to$B$\to$A$\to$C & $1.27_{\pm0.50}$ & $9.68_{\pm3.12}$ & $2.39_{\pm1.76}$ & $21.68_{\pm11.80}$ & $1.36_{\pm0.53}$ & $14.11_{\pm9.07}$ & $1.49_{\pm0.56}$ & $19.62_{\pm11.26}$ \\
D$\to$B$\to$C$\to$A & $1.14_{\pm0.18}$ & $13.15_{\pm3.24}$ & $1.10_{\pm0.16}$ & $10.39_{\pm3.88}$ & $1.18_{\pm0.23}$ & $11.06_{\pm1.64}$ & $1.36_{\pm0.45}$ & $15.95_{\pm7.94}$ \\
D$\to$C$\to$A$\to$B & $1.04_{\pm0.06}$ & $10.12_{\pm1.57}$ & $2.24_{\pm0.90}$ & $24.82_{\pm7.94}$ & $1.01_{\pm0.13}$ & $7.89_{\pm2.05}$ & $1.02_{\pm0.11}$ & $9.98_{\pm2.72}$ \\
D$\to$C$\to$B$\to$A & $1.01_{\pm0.08}$ & $9.44_{\pm2.96}$ & $1.03_{\pm0.13}$ & $9.01_{\pm2.85}$ & $1.06_{\pm0.10}$ & $11.17_{\pm2.48}$ & $1.32_{\pm0.52}$ & $14.56_{\pm7.85}$ \\
\bottomrule
\end{tabular*}
\end{table*}

\begin{table*}[!t]
\centering
\caption{Backward transfer by task order (mean $\pm$ sample standard deviation over three replicates). FDE is in metres and MR in percentage points. Bold as in Table~\ref{tab:main_results_plasticity}.}
\label{tab:main_results_stability}
\setlength{\tabcolsep}{3.5pt}
\renewcommand{\arraystretch}{1.0}
\fontsize{7}{7.7}\selectfont
\begin{tabular*}{\textwidth}{@{\extracolsep{\fill}}lcccccccc}
\toprule
Order & \multicolumn{2}{c}{Vanilla} & \multicolumn{2}{c}{DER} & \multicolumn{2}{c}{SyReM} & \multicolumn{2}{c}{GSM} \\
\cmidrule(lr){2-3}\cmidrule(lr){4-5}\cmidrule(lr){6-7}\cmidrule(lr){8-9}
 & FDE-BWT & MR-BWT & FDE-BWT & MR-BWT & FDE-BWT & MR-BWT & FDE-BWT & MR-BWT \\
\midrule
A$\to$B$\to$C$\to$D & $-16.39_{\pm1.57}$ & $-46.19_{\pm3.15}$ & $-10.99_{\pm5.05}$ & $-41.41_{\pm0.95}$ & $-14.77_{\pm8.97}$ & $-52.95_{\pm2.18}$ & $-12.00_{\pm3.63}$ & $-54.18_{\pm2.86}$ \\
A$\to$B$\to$D$\to$C & $-8.41_{\pm5.26}$ & $-42.47_{\pm2.39}$ & $-12.28_{\pm5.50}$ & $-44.11_{\pm2.93}$ & $-13.55_{\pm5.89}$ & $-54.58_{\pm3.96}$ & $-15.61_{\pm9.07}$ & $-48.23_{\pm11.43}$ \\
A$\to$C$\to$B$\to$D & $-13.97_{\pm7.56}$ & $-47.29_{\pm3.65}$ & $-13.49_{\pm6.02}$ & $-46.46_{\pm2.77}$ & $-5.96_{\pm0.70}$ & $-50.20_{\pm1.11}$ & $-12.35_{\pm2.34}$ & $-49.80_{\pm3.44}$ \\
A$\to$C$\to$D$\to$B & {\boldmath$-19.84_{\pm4.77}$} & $-42.47_{\pm11.19}$ & $-5.36_{\pm2.77}$ & $-37.45_{\pm12.64}$ & $-9.18_{\pm3.12}$ & $-53.14_{\pm5.23}$ & $-8.97_{\pm5.24}$ & $-48.22_{\pm7.97}$ \\
A$\to$D$\to$B$\to$C & $-19.42_{\pm4.94}$ & $-45.86_{\pm2.94}$ & $-5.44_{\pm1.35}$ & $-40.02_{\pm8.01}$ & $-7.15_{\pm4.53}$ & $-45.12_{\pm8.76}$ & $-18.75_{\pm5.52}$ & {\boldmath$-50.08_{\pm2.73}$} \\
A$\to$D$\to$C$\to$B & $-11.50_{\pm6.68}$ & $-33.69_{\pm21.46}$ & $-10.01_{\pm9.22}$ & $-41.18_{\pm7.96}$ & $-13.73_{\pm5.86}$ & {\boldmath$-56.97_{\pm0.56}$} & $-14.32_{\pm0.59}$ & $-49.42_{\pm2.57}$ \\
B$\to$A$\to$C$\to$D & $0.38_{\pm1.30}$ & $-4.77_{\pm2.03}$ & $0.38_{\pm0.41}$ & $2.80_{\pm7.91}$ & $-0.35_{\pm0.64}$ & $-12.22_{\pm2.50}$ & $-0.68_{\pm1.66}$ & $-8.45_{\pm10.69}$ \\
B$\to$A$\to$D$\to$C & $-0.95_{\pm0.77}$ & $-8.34_{\pm10.34}$ & $-0.13_{\pm0.46}$ & $-3.38_{\pm7.23}$ & $-0.18_{\pm1.07}$ & $-7.88_{\pm12.83}$ & $-0.36_{\pm1.37}$ & $-5.76_{\pm7.80}$ \\
B$\to$C$\to$A$\to$D & $0.17_{\pm0.84}$ & $4.50_{\pm2.97}$ & $0.57_{\pm0.80}$ & $7.14_{\pm4.99}$ & $-0.51_{\pm1.00}$ & $-4.43_{\pm9.09}$ & $-0.63_{\pm0.44}$ & $-3.01_{\pm2.01}$ \\
B$\to$C$\to$D$\to$A & $0.97_{\pm1.18}$ & $12.43_{\pm9.35}$ & $1.08_{\pm0.29}$ & $10.74_{\pm3.78}$ & $-0.47_{\pm0.65}$ & $-3.43_{\pm7.03}$ & $-0.72_{\pm0.92}$ & $1.58_{\pm3.47}$ \\
B$\to$D$\to$A$\to$C & $0.44_{\pm0.85}$ & $4.86_{\pm1.12}$ & $-0.01_{\pm0.80}$ & $5.39_{\pm4.17}$ & $-0.38_{\pm0.66}$ & $-5.72_{\pm3.49}$ & $-0.85_{\pm0.32}$ & $-2.73_{\pm2.70}$ \\
B$\to$D$\to$C$\to$A & $0.43_{\pm1.25}$ & $7.10_{\pm11.32}$ & $0.87_{\pm1.00}$ & $11.62_{\pm2.77}$ & $-1.02_{\pm1.18}$ & $-0.11_{\pm3.79}$ & $-0.81_{\pm0.70}$ & $1.27_{\pm4.80}$ \\
C$\to$A$\to$B$\to$D & $0.66_{\pm0.14}$ & $6.36_{\pm4.90}$ & $0.68_{\pm0.13}$ & $9.91_{\pm3.58}$ & {\boldmath$-0.10_{\pm0.38}$} & {\boldmath$-2.38_{\pm7.88}$} & $0.22_{\pm0.05}$ & $5.57_{\pm1.98}$ \\
C$\to$A$\to$D$\to$B & $0.38_{\pm0.15}$ & $3.60_{\pm5.30}$ & $0.45_{\pm0.14}$ & $8.91_{\pm3.94}$ & {\boldmath$0.05_{\pm0.12}$} & $1.92_{\pm3.45}$ & $0.42_{\pm0.30}$ & $8.87_{\pm6.16}$ \\
C$\to$B$\to$A$\to$D & $0.97_{\pm0.31}$ & $11.47_{\pm3.74}$ & $0.80_{\pm0.19}$ & $14.11_{\pm1.94}$ & $0.11_{\pm0.17}$ & {\boldmath$0.22_{\pm4.51}$} & $0.46_{\pm0.63}$ & $10.30_{\pm4.48}$ \\
C$\to$B$\to$D$\to$A & $1.12_{\pm0.19}$ & $13.35_{\pm3.41}$ & $1.29_{\pm0.34}$ & $15.87_{\pm2.03}$ & {\boldmath$-0.07_{\pm0.25}$} & {\boldmath$-2.14_{\pm6.78}$} & $0.26_{\pm0.37}$ & $7.52_{\pm6.05}$ \\
C$\to$D$\to$A$\to$B & $0.64_{\pm0.24}$ & $14.77_{\pm3.67}$ & $0.66_{\pm0.18}$ & $14.69_{\pm2.58}$ & {\boldmath$0.21_{\pm0.25}$} & {\boldmath$3.29_{\pm3.01}$} & $0.58_{\pm0.13}$ & $13.43_{\pm1.53}$ \\
C$\to$D$\to$B$\to$A & $0.82_{\pm0.12}$ & $16.91_{\pm1.55}$ & $0.73_{\pm0.48}$ & $13.16_{\pm8.55}$ & {\boldmath$0.21_{\pm0.15}$} & {\boldmath$4.24_{\pm3.76}$} & $0.85_{\pm0.51}$ & $15.54_{\pm6.66}$ \\
D$\to$A$\to$B$\to$C & $0.40_{\pm0.45}$ & $10.65_{\pm7.35}$ & $0.71_{\pm0.16}$ & $14.06_{\pm2.05}$ & $-0.49_{\pm1.34}$ & {\boldmath$-3.59_{\pm12.00}$} & $1.38_{\pm1.66}$ & $15.06_{\pm9.24}$ \\
D$\to$A$\to$C$\to$B & $0.45_{\pm0.07}$ & $5.37_{\pm3.07}$ & $0.64_{\pm0.18}$ & $11.91_{\pm4.44}$ & $0.29_{\pm0.38}$ & $2.62_{\pm1.80}$ & $-0.75_{\pm1.53}$ & $0.84_{\pm2.56}$ \\
D$\to$B$\to$A$\to$C & $0.86_{\pm0.49}$ & $17.83_{\pm2.77}$ & $1.04_{\pm0.34}$ & $15.53_{\pm2.89}$ & {\boldmath$0.21_{\pm0.06}$} & {\boldmath$3.21_{\pm1.64}$} & $0.90_{\pm0.16}$ & $8.44_{\pm5.08}$ \\
D$\to$B$\to$C$\to$A & $1.03_{\pm0.17}$ & $17.66_{\pm3.60}$ & $0.95_{\pm0.14}$ & $16.63_{\pm3.17}$ & {\boldmath$0.01_{\pm0.19}$} & {\boldmath$2.55_{\pm1.02}$} & $0.84_{\pm0.58}$ & $13.79_{\pm5.05}$ \\
D$\to$C$\to$A$\to$B & $0.73_{\pm0.11}$ & $15.61_{\pm1.74}$ & $0.73_{\pm0.37}$ & $12.99_{\pm5.99}$ & {\boldmath$0.21_{\pm0.22}$} & $6.13_{\pm6.09}$ & $0.54_{\pm0.25}$ & $10.47_{\pm4.67}$ \\
D$\to$C$\to$B$\to$A & $0.85_{\pm0.32}$ & $15.16_{\pm6.63}$ & $0.91_{\pm0.27}$ & $18.19_{\pm3.59}$ & {\boldmath$0.03_{\pm0.12}$} & {\boldmath$-0.22_{\pm4.14}$} & $0.53_{\pm0.41}$ & $11.18_{\pm5.78}$ \\
\bottomrule
\end{tabular*}
\end{table*}

\subsection{The Starting Task and the Cold Start of Task A}
\label{subsec:starting_task}
Table~\ref{tab:bwt_diagnostic} resolves the FDE summaries by starting task. When task A is learned first, the first-stage error $R_{1,1}$ averages 21.52~m over the 72 runs, ranging from 4.7 to 42.9~m, with a first-stage miss rate of 93.8\%. For the other starting tasks, $R_{1,1}$ averages 1.04 to 2.77~m. When A is learned at positions 2, 3, and 4 instead, its own stage error is 3.30, 1.87, and 1.18~m, against 0.88 to 1.32~m for the other tasks at warm positions, and its final error is 0.96 to 1.18~m in every position. The algorithm-averaged FDE-CT accordingly falls from 6.25~m with A in position 1 to 1.76, 1.42, and 1.20~m with A in positions 2, 3, and 4. Task A is both the smallest task and the most distinct in speed, and its failure in the first stage may stem from either property or both. The algorithms already differ within this stage: DER and SyReM reach lower A-first first-stage errors than Vanilla and GSM (17.6 and 18.5~m against 26.4 and 23.6~m), consistent with their memory losses acting while the buffer fills.

Removing the first-stage term changes the picture. For the warm-start error $\mathrm{CT}_{2:N}$, the order component drops to 25.6\% (FDE) and 23.8\% (MR), the starting-task share to 4.9\% and 7.9\%, and within-cell training variation rises to 52.0\% and 49.2\% (Table~\ref{tab:decomposition}). A-first orders are no longer the worst starting group (Table~\ref{tab:bwt_diagnostic}), but only because their $\mathrm{CT}_{2:N}$ averages tasks B, C, and D, whereas for the other starting tasks it also includes the warm-position error of A, about twice that of the other tasks. Without that term, A-first orders have the highest warm-start error (1.16~m against 0.95 to 1.05~m). The residual order effect is thus comparable to training variation and is organised by the position of A, whose group means account for 13.3\% (FDE) and 15.7\% (MR) of the total sum of squares, half to two-thirds of the residual order term.
\begin{table}[!t]
\centering
\caption{FDE (m) by starting task over 72 runs per row: initial error $R_{1,1}$, CT with and without (CT$_{2:N}$) the first stage, subsequent error $P$, reference $B$, BWT $=P-B$, and final error.}
\label{tab:bwt_diagnostic}
\setlength{\tabcolsep}{2.5pt}
\footnotesize
\begin{tabular*}{\columnwidth}{@{\extracolsep{\fill}}lrrrrrrr@{}}
\toprule
Start & $R_{1,1}$ & CT & CT$_{2:N}$ & $P$ & $B$ & BWT & Final \\
\midrule
A & $21.52$ & $6.25$ & $1.16$ & $1.46$ & $13.69$ & $-12.23$ & $1.28$ \\
B & $2.77$ & $1.81$ & $1.48$ & $2.34$ & $2.46$ & $-0.12$ & $1.52$ \\
C & $1.04$ & $1.12$ & $1.15$ & $1.64$ & $1.13$ & $0.52$ & $1.50$ \\
D & $1.32$ & $1.45$ & $1.50$ & $2.04$ & $1.50$ & $0.54$ & $1.48$ \\
\bottomrule
\end{tabular*}
\end{table}

\subsection{Descriptor Associations and the Position of Task A}
\label{subsubsec:dvscl}
Table~\ref{tab:correlation_comparison} shows how the associations of $D$ with CL performance change across the five candidate task distances. W1 on speed, the primary distance, was selected against these same outcomes in the exploratory scan (Appendix~\ref{subsec:appendix_scan}). MMD and W1 on speed order the descriptor values identically and therefore give identical Spearman correlations. Under the acceleration distance, the pair containing A is the closer pair in two of the three partitions, so $D$ changes sign for the 16 orders built on those partitions and no longer tracks the position of A (Spearman $-0.61$ against $+0.79$ for speed). The correlations in the last column reverse in sign accordingly, with smaller magnitudes for BWT.

\begin{table}[!t]
\centering
\caption{Spearman correlations between candidate transition-direction descriptors and order-level performance. W1: Wasserstein-1. Bold: $|\rho|\geq0.7$.}
\label{tab:correlation_comparison}
\setlength{\tabcolsep}{2.6pt}
\renewcommand{\arraystretch}{1.02}
\footnotesize
\begin{tabular*}{\columnwidth}{@{\extracolsep{\fill}}llrrrrr@{}}
\toprule
Algorithm & Metric & \shortstack{W1\\speed} & \shortstack{KL\\speed} & \shortstack{JS\\speed} & \shortstack{MMD\\speed} & \shortstack{W1\\accel.} \\
\midrule
Vanilla & FDE-CT & {\boldmath$-0.75$} & $-0.23$ & $-0.20$ & {\boldmath$-0.75$} & $+0.69$ \\
 & MR-CT & {\boldmath$-0.75$} & $-0.09$ & $-0.12$ & {\boldmath$-0.75$} & $+0.69$ \\
 & FDE-BWT & $+0.66$ & $+0.13$ & $+0.10$ & $+0.66$ & $-0.55$ \\
 & MR-BWT & {\boldmath$+0.81$} & $+0.26$ & $+0.32$ & {\boldmath$+0.81$} & $-0.46$ \\
\midrule
DER & FDE-CT & $-0.57$ & $-0.29$ & $-0.36$ & $-0.57$ & $+0.46$ \\
 & MR-CT & $-0.54$ & $-0.22$ & $-0.27$ & $-0.54$ & $+0.48$ \\
 & FDE-BWT & $+0.67$ & $+0.10$ & $+0.12$ & $+0.67$ & $-0.52$ \\
 & MR-BWT & {\boldmath$+0.74$} & $+0.28$ & $+0.32$ & {\boldmath$+0.74$} & $-0.43$ \\
\midrule
SyReM & FDE-CT & $-0.67$ & $-0.29$ & $-0.26$ & $-0.67$ & {\boldmath$+0.74$} \\
 & MR-CT & {\boldmath$-0.72$} & $-0.22$ & $-0.23$ & {\boldmath$-0.72$} & {\boldmath$+0.76$} \\
 & FDE-BWT & $+0.55$ & $+0.31$ & $+0.36$ & $+0.55$ & $-0.35$ \\
 & MR-BWT & {\boldmath$+0.72$} & $+0.35$ & $+0.43$ & {\boldmath$+0.72$} & $-0.42$ \\
\midrule
GSM & FDE-CT & $-0.57$ & $-0.30$ & $-0.29$ & $-0.57$ & $+0.66$ \\
 & MR-CT & $-0.53$ & $-0.23$ & $-0.23$ & $-0.53$ & $+0.69$ \\
 & FDE-BWT & $+0.55$ & $+0.32$ & $+0.30$ & $+0.55$ & $-0.37$ \\
 & MR-BWT & $+0.69$ & $+0.40$ & $+0.44$ & $+0.69$ & $-0.43$ \\
\midrule
\multicolumn{2}{l}{Mean $|\rho|$} & $0.66$ & $0.25$ & $0.27$ & $0.66$ & $0.54$ \\
\bottomrule
\end{tabular*}
\end{table}

\begin{figure*}[!t]
\centering
\includegraphics[width=\textwidth]{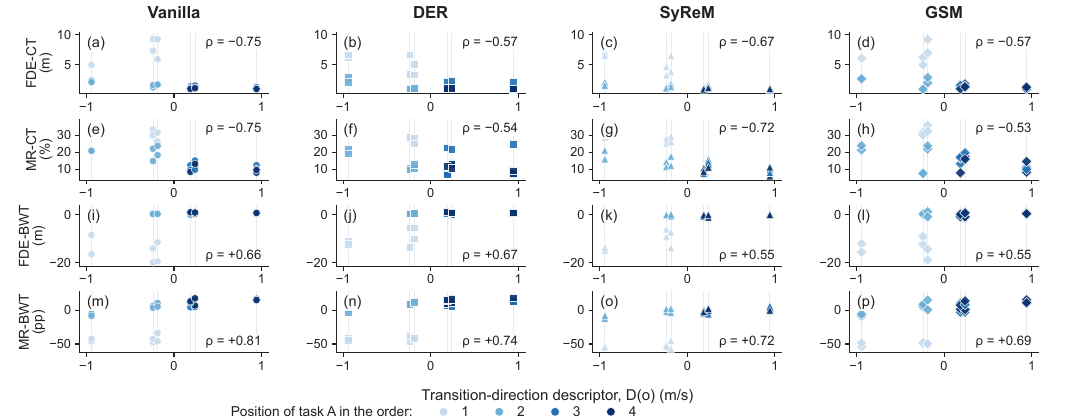}
\caption{Transition-direction descriptor versus performance for each algorithm. Points are replicate means for individual orders, coloured by the position of task A, and $\rho$ is the Spearman correlation with $D$. Grey guides mark the six distinct descriptor values.}
\label{fig:unified_D_main}
\end{figure*}
Across the 16 algorithm--metric series, absolute Spearman correlations with $D$ range from 0.53 to 0.81, with algorithm-averaged values of $-0.64$, $-0.63$, $+0.61$, and $+0.74$ for FDE-CT, MR-CT, FDE-BWT, and MR-BWT. Fig.~\ref{fig:unified_D_main} resolves these associations for each algorithm and metric across the six tied descriptor levels, and the two clusters visible in every panel coincide with A in positions 1--2 (negative $D$) and 3--4 (positive $D$).

Table~\ref{tab:starting_correlations} quantifies this reading. The position of A alone correlates more strongly than $D$ with each of the four primary metrics ($-0.77$, $-0.79$, $+0.73$, and $+0.77$). Excluding A-first orders reduces the mean association with $D$ in each of these metrics, removing starting-task means leaves $-0.48$, $-0.49$, $+0.29$, and $+0.62$, and removing position-of-A means leaves $-0.23$, $-0.11$, $+0.22$, and $+0.53$. For the warm-start error the marginal associations are weaker ($-0.37$ and $-0.27$ with $D$, and $-0.20$ and $-0.10$ with the position of A), and excluding A-first orders strengthens them because these orders combine the most negative $D$ with low warm-start error. $D$ therefore acts mainly as an index of the position of task A, the most distinct and smallest task. Once that position is removed, a substantial association remains only for MR-BWT ($+0.53$).
\begin{table}[!t]
\centering
\caption{Mean Spearman associations across algorithms. $\rho(D)$: descriptor and order means. Excl.\ A-first: 18 orders. Start-adj.: rank residuals after removing starting-task means. $\rho(p_A)$: position of task A. $p_A$-adj.: rank residuals after removing position-of-A means.}
\label{tab:starting_correlations}
\setlength{\tabcolsep}{2.5pt}
\footnotesize
\begin{tabular*}{\columnwidth}{@{\extracolsep{\fill}}lrrrrr@{}}
\toprule
Metric & $\rho(D)$ & Excl.\ A-first & Start-adj. & $\rho(p_A)$ & $p_A$-adj. \\
\midrule
FDE-CT & $-0.64$ & $-0.47$ & $-0.48$ & $-0.77$ & $-0.23$ \\
MR-CT & $-0.63$ & $-0.43$ & $-0.49$ & $-0.79$ & $-0.11$ \\
FDE-BWT & $+0.61$ & $+0.40$ & $+0.29$ & $+0.73$ & $+0.22$ \\
MR-BWT & $+0.74$ & $+0.64$ & $+0.62$ & $+0.77$ & $+0.53$ \\
\midrule
FDE-CT$_{2:N}$ & $-0.37$ & $-0.49$ & $-0.58$ & $-0.20$ & $-0.27$ \\
MR-CT$_{2:N}$ & $-0.27$ & $-0.47$ & $-0.50$ & $-0.10$ & $-0.20$ \\
\bottomrule
\end{tabular*}
\end{table}

\subsection{Backward Transfer and the Initial Reference}
\label{subsec:bwt_audit}
The large negative BWT values for A-first orders motivate a decomposition of the initial reference. Expanding~\eqref{eq:BWT} gives
\begin{equation}
\mathrm{BWT}=P-B,
\label{eq:bwt_decomp}
\end{equation}
where
\begin{align}
P&=\frac1{N-1}\sum_{r=2}^N\frac1{r-1}\sum_{i<r}R_{r,i},\label{eq:P}\\
B&=\frac1{N-1}\sum_{r=2}^N\frac1{r-1}\sum_{i<r}R_{i,i}.\label{eq:B}
\end{align}
Both $P$ and $B$ are error levels: the average subsequent error on previously seen tasks and the corresponding weighted initial reference. A high $B$ makes BWT very negative even when $P$ is moderate. The first task is referenced at every later stage, so for $N=4$ its coefficient in $B$ is $11/18$, compared with $5/18$ and $1/9$ for the second and third tasks.

For A-first runs, the subsequent-error and reference terms are 1.46 and 13.69~m, yielding BWT $=-12.23$~m with a final mean error of 1.28~m (Table~\ref{tab:bwt_diagnostic}). Across algorithms, the correlation of $D$ with the FDE reference term $B$ is between $-0.69$ and $-0.57$, whereas its correlation with the subsequent-error term $P$ is only $+0.02$ to $+0.19$. The association of $D$ with FDE-BWT therefore comes mainly from the order-dependent reference term $B$ rather than from the subsequent error $P$.

\begin{table}[!t]
\centering
\caption{Final errors: mean [minimum, maximum] over order-level replicate means, and the bootstrap probability (Prob.) of the smallest uniform mean. FDE is in metres and MR in percent. Bold: smallest mean.}
\label{tab:final_error}
\setlength{\tabcolsep}{3pt}
\footnotesize
\begin{tabular*}{\columnwidth}{@{\extracolsep{\fill}}lcrcr@{}}
\toprule
Algorithm & FDE: mean [min, max] & Prob. & MR: mean [min, max] & Prob. \\
\midrule
Vanilla & 1.63 [1.18, 2.14] & $0.00$ & 17.22 [10.84, 25.67] & $0.00$ \\
DER & 1.63 [1.17, 2.34] & $0.00$ & 17.80 [11.47, 29.09] & $0.00$ \\
SyReM & \textbf{1.08 [0.87, 1.99]} & $1.00$ & \textbf{9.23 [4.38, 24.69]} & $1.00$ \\
GSM & 1.43 [1.03, 2.44] & $0.00$ & 15.86 [8.05, 36.25] & $0.00$ \\
\bottomrule
\end{tabular*}
\end{table}

Final error gives a different comparison (Table~\ref{tab:final_error}). SyReM has the lowest uniform-order final FDE (1.08~m) and MR (9.23\%) with bootstrap probability 1.00 for both, and its per-order lead is separable in 17 and 16 of the 24 orders. GSM's smallest uniform-order FDE-BWT ($-3.33$~m), by contrast, is within 0.03~m of Vanilla, with probabilities 0.51 and 0.46 (Table~\ref{tab:weighted_metrics}), and this lead comes entirely from the six A-first orders, where the first-stage reference is largest. Over the 18 remaining orders SyReM has the smallest score in all four metrics, with bootstrap probabilities of 0.90, 0.89, 0.98, and 1.00. Improvement relative to an algorithm-specific initial checkpoint and absolute final prediction quality thus lead to different comparisons, and the final-error comparison is the one the replicates resolve. Because CT and BWT share the initial errors $R_{i,i}$ with opposite signs, a high initial error raises CT and lowers BWT, which accounts for part of their opposite associations with $D$.

\subsection{Different Paths from the Same Starting Task}
\label{subsubsec:max_min_paths}
\begin{figure*}[!t]
\centering
\includegraphics[width=\textwidth]{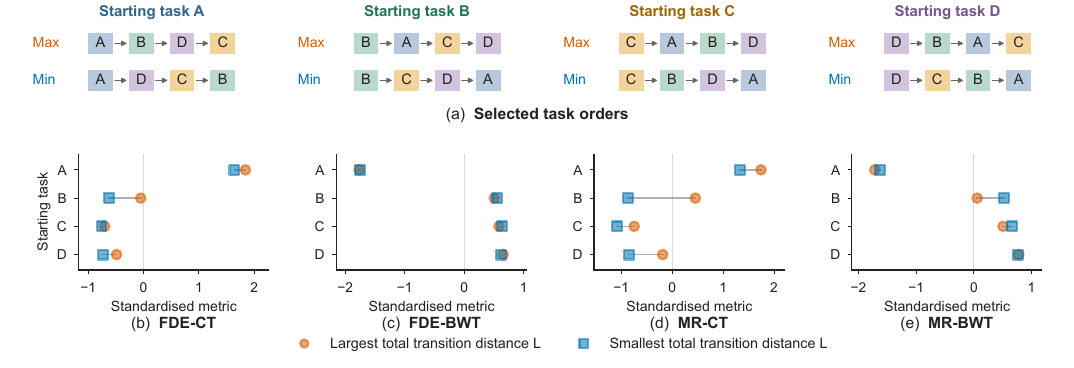}
\caption{Paths with the largest and smallest total transition distance $L$ for each starting task. Order strips identify the paths, and paired points compare algorithm-averaged scores, standardised separately for each metric across task orders.}
\label{fig:max_min_paths}
\end{figure*}

Fixing the first task leaves six orders of the remaining three. Fig.~\ref{fig:max_min_paths} compares the largest- and smallest-$L$ paths within each group under the W1 matrix on speed. For every starting task the maximum-distance path has the smaller $D$. The paired paths differ in measured performance, with a direction that depends on the metric. The two B-first paths, for example, have algorithm-averaged FDE-CT values of 2.55 and 1.33~m. Order thus matters beyond the starting task: in the B-, C-, and D-first groups the maximum-distance path also places A earlier (position 2, 2, and 3 against 4) and has the higher current-task error in both FDE and MR, consistent with the position-of-A pattern of the residual order effect in Section~\ref{subsec:starting_task}.

\subsection{Effect of the Order Policy}
\label{subsec:policy_effect}
\begin{table}[!t]
\centering
\caption{Policy scores (units as in Tables~\ref{tab:main_results_plasticity} and~\ref{tab:main_results_stability}) under the uniform policy, the descriptor-weighted policy, and the uniform policy over the 18 orders that do not start with A, with the bootstrap probability of the smallest score under that policy in parentheses. Bold: smallest score per row.}
\label{tab:weighted_metrics}
\setlength{\tabcolsep}{1.5pt}
\scriptsize
\begin{tabular*}{\columnwidth}{@{\extracolsep{\fill}}llrrrr@{}}
\toprule
Metric & Policy & Vanilla & DER & SyReM & GSM \\
\midrule
FDE-CT & Uniform & $2.94$ (0.00) & $2.42$ (0.29) & {\boldmath$2.33$} (0.71) & $2.94$ (0.00) \\
 & $D$-weighted & $3.76$ (0.00) & $3.06$ (0.43) & {\boldmath$3.01$} (0.57) & $3.76$ (0.00) \\
 & Excl.\ A-first & $1.43$ (0.06) & $1.48$ (0.04) & {\boldmath$1.31$} (0.90) & $1.62$ (0.00) \\
\midrule
MR-CT & Uniform & $18.10$ (0.00) & $17.30$ (0.05) & {\boldmath$16.04$} (0.94) & $19.58$ (0.00) \\
 & $D$-weighted & $21.87$ (0.00) & $20.12$ (0.21) & {\boldmath$19.34$} (0.79) & $22.98$ (0.00) \\
 & Excl.\ A-first & $13.86$ (0.01) & $13.32$ (0.09) & {\boldmath$11.98$} (0.89) & $15.19$ (0.00) \\
\midrule
FDE-BWT & Uniform & $-3.30$ (0.46) & $-1.88$ (0.00) & $-2.77$ (0.03) & {\boldmath$-3.33$} (0.51) \\
 & $D$-weighted & $-1.46$ (0.36) & $-0.43$ (0.00) & $-1.23$ (0.02) & {\boldmath$-1.55$} (0.61) \\
 & Excl.\ A-first & $0.57$ (0.00) & $0.69$ (0.00) & {\boldmath$-0.13$} (0.98) & $0.12$ (0.02) \\
\midrule
MR-BWT & Uniform & $-3.89$ (0.00) & $-2.10$ (0.00) & {\boldmath$-13.79$} (1.00) & $-8.17$ (0.00) \\
 & $D$-weighted & $5.17$ (0.00) & $6.01$ (0.00) & {\boldmath$-6.04$} (1.00) & $0.74$ (0.00) \\
 & Excl.\ A-first & $9.14$ (0.00) & $11.13$ (0.00) & {\boldmath$-1.00$} (1.00) & $5.77$ (0.00) \\
\bottomrule
\end{tabular*}
\end{table}

Table~\ref{tab:weighted_metrics} compares the uniform and descriptor-weighted scores. Weighting increases the reported CT error for every algorithm: SyReM's FDE-CT changes from 2.33 to 3.01~m and Vanilla's from 2.94 to 3.76~m, and BWT shifts upward, for example from $-13.79$ to $-6.04$ percentage points for SyReM's MR-BWT. These shifts follow from the weighting directions: 74.2\% of the CT weight falls on the 12 orders with A in positions 1--2 and 37.1\% on the six A-first orders, against 50\% and 25\% under the uniform policy, so the policy emphasises orders that place A early. The effective number of weighted orders is 17.54 of 24.

The complete four-algorithm ranking is identical under the two policies in all four metrics. The bootstrap probabilities identify the leaders that the replicates resolve, SyReM for MR-CT (0.94) and MR-BWT (1.00), while the FDE-CT leader (SyReM, 0.71) and the FDE-BWT leader (GSM, 0.51) remain close to their runners-up. The descriptor weighting thus shifts the aggregate scores and leaves the selected algorithm unchanged in every metric.

\section{Discussion}
\label{sec:discussion}

\subsection{Interpreting the Influence of Task Order}
\label{subsec:disc_order}
Across the 24 orders, task order matters chiefly through the learning quality of the first task. Starting-task identity explains 90\% to 99\% of the order term in each primary metric, so the across-order component of~\eqref{eq:variation_axes} is essentially a starting-task component, driven by the cold start of the smallest task. Beyond the first stage, the order effect becomes comparable to training variability and is organised by where task A is placed: the smallest task stays harder to learn at later positions, and its position accounts for half to two-thirds of the residual order term. Within the first stage, DER and SyReM already reach lower cold-start errors than Vanilla and GSM.

The transition-direction descriptor inherits this structure. Because the pair containing task A is the more distant pair in every partition of the task set, $D$ tracks the position of task A, and that position correlates more strongly with every primary metric than $D$ does.

The metric decomposition explains why the same first stage moves CT and BWT in opposite directions: the cold-start error enters CT with weight $1/N$, whereas in the reference term of $\mathrm{BWT}=P-B$ the first task carries coefficient $11/18$ for $N=4$. For A-first orders the reference term is an order of magnitude larger than the subsequent error, so the strongly negative BWT records recovery from a poor start, and the order dependence of BWT is largely a reference-term effect.

\subsection{Implications for Algorithm Comparison and Benchmark Design}
\label{subsec:disc_implications}
Four reporting items follow for continual learning benchmarks.
\begin{enumerate}
\item State the order distribution behind every aggregate score, the policy $q$ in~\eqref{eq:population_policy}.
\item Report order-resolved and starting-task-resolved outcomes, since changing the order can move a score by far more than changing the algorithm (Tables~\ref{tab:main_results_plasticity} and~\ref{tab:bwt_diagnostic}).
\item Report the warm-start error $\mathrm{CT}_{2:N}$ and the final error alongside CT and BWT.
\item Fix any descriptor and weighting rule before the outcomes are seen, so that the policy expresses an intended emphasis. Here, the descriptor-weighted policy took its weighting directions from the present outcomes, put most of its CT weight on orders with task A in the first two positions, and left the ranking unchanged (Table~\ref{tab:weighted_metrics}).
\end{enumerate}

For the compared algorithms, the clearest result is that SyReM attains the smallest final error and the smallest MR-BWT with bootstrap probability 1.00, leads MR-CT with probability 0.94, and leads all four metrics once the A-first orders are set aside. Final error also separates per-order leaders far more often than CT and BWT (16 or 17 of 24 orders against 2 to 11), and every separable final-error leader is SyReM, so final error is the comparison these replicates resolve most clearly. The interaction term exceeds its within-cell noise level for FDE-CT and FDE-BWT, indicating that algorithms differ in their sensitivity to order. For example, the FDE-CT range across orders is 5.8~m for DER and 6.3~m for SyReM, against 8.2~m for GSM and 8.3~m for Vanilla (Table~\ref{tab:main_results_plasticity}).

\subsection{Extension to Other Task Sets}
\label{subsec:disc_extension}
The protocol applies to any task set, whereas the magnitudes reported here belong to one benchmark with four tasks, one backbone, and three replicates per cell. For larger $N$, sampling from a declared order distribution replaces enumeration, and repeated samples separate coverage uncertainty from training variability. Samples should balance starting tasks, which carried most of the order term here, and the warm-start decomposition should accompany CT. Pre-specifying the descriptor and weights turns the descriptor-weighted policy from the exploratory form used here into a confirmatory one. Task representations beyond vehicle speed, such as distances on the inputs or on backbone features, would extend the protocol to domains without an application-specific variable, and independently constructed task sets will show how far the starting-task effect generalises beyond motion forecasting.

\section{Conclusion}
\label{sec:conclusion}
This study formulates continual learning evaluation as an expectation over a declared distribution of task orders and applies it to motion forecasting across all 24 orders of four tasks, four algorithms, and three replicates. Task order accounts for 65\% to 91\% of the variation in current-task error and backward transfer, and almost all of this effect comes from the identity of the starting task, acting through the cold start of the smallest task. Beyond the first stage, the order effect is comparable to training variability, and most rank reversals between orders lie within it. The first task dominates the reference term of backward transfer, so relative improvement and absolute final error identify different leading algorithms for displacement error, and the transition-direction descriptor tracks where the smallest task is placed, leaving the ranking unchanged under descriptor weighting. Benchmarks that declare their order distribution and report order-resolved, starting-task-resolved, warm-start, and final errors make such effects visible and their algorithm comparisons interpretable.

\appendices
\section{Analysis Details and Reproducibility}
\label{Appendix-Params}
\subsection{Feature and Distance Reconstruction}
\label{subsec:appendix_distance}
Target-agent speed and acceleration are computed from the trajectory arrays of all 12 location subsets. Feature counts and source hashes accompany the released inputs. Equation~\eqref{eq:symmetry} gives the W1 distance. For KL and JS, let $c_{ib}$ count speed observations in bin $b$ of $H=100$ shared equal-width bins spanning the pooled range. With $\epsilon=10^{-10}$, the smoothed probabilities are
\begin{equation}
p_{ib}=\frac{c_{ib}/\sum_{h=1}^{H}c_{ih}+\epsilon}{1+H\epsilon},
\label{eq:histogram_probabilities}
\end{equation}
and, writing $m_{ijb}=(p_{ib}+p_{jb})/2$,
\begin{align}
d^{\rm KL}_{ij}&=\sum_{b=1}^{H}p_{ib}\log\frac{p_{ib}}{p_{jb}},\label{eq:kl}\\
d^{\rm JS}_{ij}&=\frac12\sum_{b=1}^{H}\left(p_{ib}\log\frac{p_{ib}}{m_{ijb}}
+p_{jb}\log\frac{p_{jb}}{m_{ijb}}\right),
\label{eq:js}
\end{align}
with natural logarithms. JS denotes the divergence rather than its square root. For MMD, a fixed sample of $n=2{,}000$ speed observations per task is standardised with the pooled mean and population standard deviation of all retained speed observations, $z_{iu}=(v_{iu}-\mu_v)/\sigma_v$, and compared with the kernel $k(z,z')=\exp[-(z-z')^2]$. The same samples are reused across pairs, and $d^{\rm MMD}_{ij}=\sqrt{\max(0,Q_{ij})}$, where $Q_{ij}$ is the biased squared discrepancy $n^{-2}\sum_{u,v}[k(z_{iu},z_{iv})+k(z_{ju},z_{jv})-2k(z_{iu},z_{jv})]$ including diagonal terms. The clipping removes roundoff below zero. MMD samples are drawn without replacement using NumPy PCG64 with seed 0, proceeding through tasks A--D in order. Exact bin edges, standardisation constants, and sampled indices accompany the data.

\subsection{Exploratory Descriptor Scan}
\label{subsec:appendix_scan}
Before the five candidates of Section~\ref{subsec:dist_shift} were fixed, an exploratory scan compared 20 distance representations (W1, KL, JS, and MMD on speed, acceleration, relative distance, and relative speed, plus four multivariate distances) combined with 15 aggregation operators over the three transitions (the individual transitions, their total, mean, minimum, maximum, range, standard deviation, slope, and six further shape summaries) against the same 24-order outcomes, giving 300 candidates. The complete scan table is included with the released inputs. The W1-on-speed slope was retained for its interpretable physical scale and consistent sign pattern across the 16 algorithm--metric series.

\subsection{Decomposition, Adjustment, and Bootstrap}
\label{subsec:appendix_decomp}
For a fixed metric, let $Z_{oas}$ be a run outcome with $K=|\mathcal O|$ orders, $J=|\mathcal A|$ algorithms, and $S$ replicates per combination. The balanced two-way decomposition is $SS_T=SS_O+SS_A+SS_{OA}+SS_W$, with $SS_O=JS\sum_o(\bar Z_{o\cdot\cdot}-\bar Z)^2$, $SS_A=KS\sum_a(\bar Z_{\cdot a\cdot}-\bar Z)^2$, $SS_{OA}=S\sum_{o,a}(\bar Z_{oa\cdot}-\bar Z_{o\cdot\cdot}-\bar Z_{\cdot a\cdot}+\bar Z)^2$, and $SS_W=\sum_{o,a,s}(Z_{oas}-\bar Z_{oa\cdot})^2$, where dots denote averaging and $\bar Z$ is the grand mean. Writing $\bar Z_{g(o)}$ for the mean over orders with the same starting task as $o$, the order term splits as $SS_O=JS\sum_o(\bar Z_{g(o)}-\bar Z)^2+JS\sum_o(\bar Z_{o\cdot\cdot}-\bar Z_{g(o)})^2$. Under within-cell noise alone, a term with $\nu$ degrees of freedom is expected to take $\nu/\nu_W$ of the within-cell share, with $\nu_W=KJ(S-1)$ and $\nu=(K-1)(J-1)$ for the interaction.

Conditional rank association for an algorithm--metric series assigns average ranks to $D$ and to the order mean, subtracts the respective mean rank within each group, and correlates the two residual vectors. The groups are the four starting tasks or the four positions of task A. The bootstrap resamples the three replicates of every order--algorithm cell with replacement (5,000 draws, NumPy PCG64 seed 0), recomputes replicate means, uniform and weighted scores, and per-order minima, and reports the fraction of draws in which each algorithm attains the smallest score together with 2.5th and 97.5th percentiles of its per-order win count. Rescaling the resampled deviations by $\sqrt{S/(S-1)}$, which corrects the plug-in variance of a three-replicate mean, changes the reported probabilities by at most 0.05. All weighted means are convex combinations of order means, and the effective number of weighted orders, $(\sum_o w_o)^2/\sum_o w_o^2$, measures weight concentration.

\ifanonymous\IEEEtriggeratref{34}\else\IEEEtriggeratref{32}\fi
\bibliographystyle{IEEEtran}
\bibliography{ref}

\end{document}